\documentclass[sigconf, nonacm]{acmart}

\usepackage{booktabs}
\usepackage{graphicx}
\usepackage{amsmath}
\usepackage{xcolor}
\usepackage{flafter}

\begin{document}

\title{Ada-MK: Adaptive MegaKernel Optimization via Automated DAG-based Search for LLM Inference}

\author{Wenxin Dong, Mingqing Hu, Guanghui Yu, Qiang Fu, Peng Xu, Hui Xu, Yue Xing, Xuewu Jiao$^\dagger$, Shuanglong Li, Lin Liu}
\affiliation{%
  \institution{Baidu Inc.}
  \city{Beijing}
  \country{China}
}

\renewcommand{\shortauthors}{Dong et al.}

\begin{abstract}
When large language models (LLMs) serve real-time inference in commercial online advertising systems, end-to-end latency must be strictly bounded to the millisecond range.  Yet every token generated during the decode phase triggers thousands of kernel launches, and kernel launch overhead alone can account for 14.6\% of end-to-end inference time.  MegaKernel eliminates launch overhead and inter-operator HBM round-trips by fusing multiple operators into a single persistent kernel.  However, existing MegaKernel implementations face a fundamental tension between portability and efficiency on resource-constrained GPUs such as NVIDIA Ada: hand-tuned solutions are tightly coupled to specific architectures and lack portability, while auto-compiled approaches introduce runtime dynamic scheduling whose branch penalties are unacceptable in latency-critical settings.

We observe that under a fixed deployment configuration, the optimal execution path of a MegaKernel is uniquely determined, and runtime dynamic decision-making can be entirely hoisted to compile time.  Building on this insight, we propose \textbf{Ada-MK}: (1)~a three-dimensional shared-memory constraint model combined with K-dimension splitting that reduces peak shared memory usage by 50\%; (2)~MLIR-based fine-grained DAG offline search that solidifies the optimal execution path, completely eliminating runtime branching; and (3)~a heterogeneous hybrid inference engine that embeds MegaKernel as a plugin into TensorRT-LLM, combining high-throughput Prefill with low-latency Decode.  On an NVIDIA L20, Ada-MK improves single-batch throughput by up to 23.6\% over vanilla TensorRT-LLM and 50.2\% over vLLM, achieving positive gains across all tested scenarios---the first industrial deployment of MegaKernel in a commercial online advertising system.
\end{abstract}


\maketitle

\section{Introduction}
\label{sec:intro}

Large language models (LLMs) are increasingly deployed in search, recommendation, and conversational applications, with parameter counts growing from billions to hundreds of billions.  Inference latency has thus become a key bottleneck for industrial adoption, especially in latency-sensitive online services where efficient inference under limited hardware resources is critical.  For extremely latency-sensitive scenarios, frequent global memory accesses and kernel launch overhead on GPUs have become the primary bottlenecks for end-to-end performance~\cite{vllm,trtllm}.  Profiling Qwen2.5-1.5B with Nsight Systems under TensorRT-LLM reveals that kernel launch overhead accounts for approximately 14.6\% of end-to-end inference time (1{,}655{,}550 launches consuming $\sim$3.3\,s).  In conventional pipelines, adjacent kernels must exchange intermediate results through HBM; MegaKernel instead leverages shared memory and registers to achieve seamless operator chaining and deep parallelism, fundamentally eliminating operator switching overhead and HBM round-trip latency.  MegaKernel achieves single-launch persistent computation through Persistent Kernels, constructs producer--consumer pipelines via Warp Specialization, and overlaps computation with memory access through TMA and asynchronous I/O~\cite{megakernel}.

Our target deployment is a commercial online advertising system running on NVIDIA Ada (L20) GPUs~\cite{ada_arch}.  Deploying MegaKernel on Ada faces severe dual constraints.  First, the business demands deterministic end-to-end latency strictly within 1--5\,ms.  Second, the Ada architecture natively lacks TMA hardware support, requiring PTX assembly and hand-crafted software pipelines to emulate asynchronous data movement; its on-chip shared memory is only half that of the H100 (128\,KB vs.\ 227\,KB)~\cite{h100_arch}, severely compressing the optimization space for pipeline stages and tile sizes.  In contrast, Hopper/Blackwell architectures provide TMA hardware support and larger shared memory (227\,KB), making MegaKernel deployment far more straightforward~\cite{megakernel,h100_arch}.  In practice, the shared memory limitation on Ada reduces the achievable pipeline stages from the theoretical optimum of 4 to only 2, incurring a pipeline duty-cycle loss exceeding 30\%.

Existing solutions fail to meet these requirements.  Stanford's MegaKernel~\cite{megakernel} delivers strong performance, but its codebase is deeply tied to Hopper/Blackwell-specific assembly optimizations and supports only a few model architectures (e.g., Llama-1B), lacking support for Qwen and other widely-used models~\cite{qwen25,qwen3}; it also lacks support for long-sequence and large-batch Prefill phases.  Mirage MPK~\cite{mpk,mirage} improves usability through auto-tuning, but its Managed Pointer mechanism introduces runtime if-else branching based on shared memory page states, degrading instruction issue efficiency in ultra-low-latency scenarios and falling short of hand-tuned operator performance.

To address these challenges, we propose Ada-MK with the following core contributions:

\begin{itemize}
  \item \textbf{Adaptive shared memory management.} We model shared memory allocation from three dimensions---hardware specifications, model architecture, and dynamic workload---and reduce peak shared memory demand by 50\% through K-dimension fine-grained splitting, while enabling cross-operator page reuse to reconstruct efficient pipelines on Ada's constrained resources.

  \item \textbf{Fine-grained DAG-based automatic search.} We leverage MLIR Lowering~\cite{mlir} to construct PTX-level dependency DAGs and solidify the optimal execution trace through offline profiling, completely eliminating runtime dynamic decision overhead.  Unlike Ansor~\cite{ansor} and other traditional auto-tuning frameworks, Ada-MK's DAG-level search captures finer-grained parallelism opportunities.

  \item \textbf{Heterogeneous hybrid inference engine.} We embed MegaKernel as a plugin into TensorRT-LLM~\cite{trtllm}, reusing TensorRT-LLM's native operators for Prefill and switching to the MegaKernel engine for Decode, achieving both high throughput and low latency---the first industrial deployment of MegaKernel.
\end{itemize}

Experiments show that Ada-MK reduces end-to-end latency by 10\%--50\% over vLLM~\cite{vllm}, SGLang~\cite{sglang}, and vanilla TensorRT-LLM~\cite{trtllm} across Qwen model series, and has been deployed in production within Baidu's commercial online advertising system.  The remainder of this paper is organized as follows: \S\ref{sec:background} presents background and related work; \S\ref{sec:arch} describes the Ada-MK overall architecture; \S\ref{sec:optimization} details the three core optimizations; \S\ref{sec:experiment} presents experimental evaluation; \S\ref{sec:conclusion} concludes with future directions.

\section{Background and Related Work}
\label{sec:background}

\begin{figure*}[!t]
  \centering
  \includegraphics[width=\textwidth]{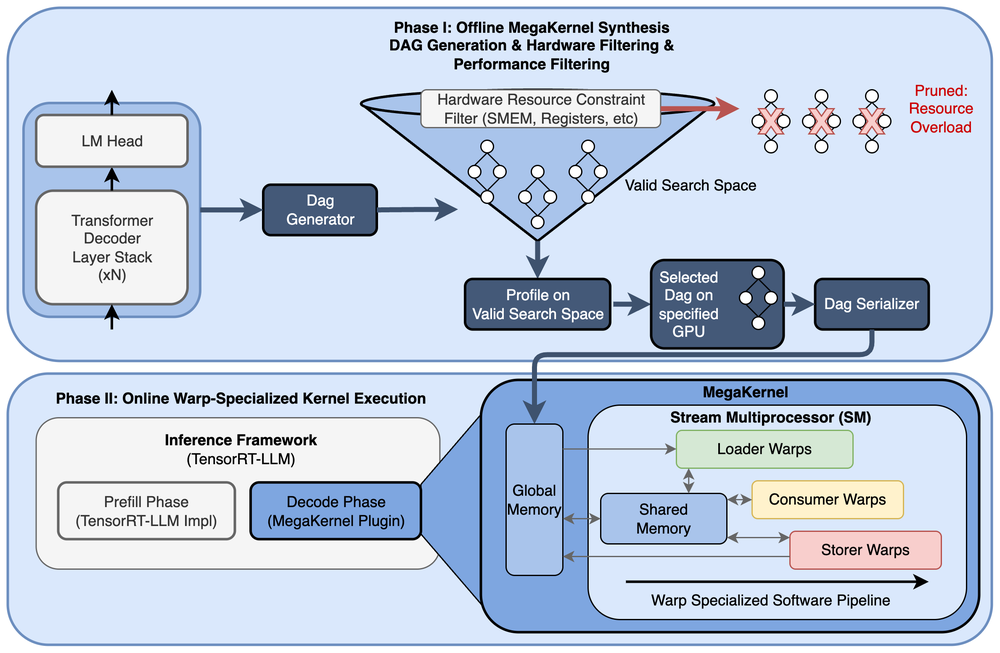}
  \caption{Ada-MK overall architecture. Phase~I (Offline MegaKernel Synthesis): the Transformer Decoder and LM Head are parsed into a fine-grained DAG, pruned by hardware resource constraints, and profiled to select the optimal execution trace, which is then serialized. Phase~II (Online Warp-Specialized Execution): the serialized MegaKernel is embedded as a TensorRT-LLM plugin, with warp-specialized roles (Loader, Consumer, Storer, Controller, Launcher) collaborating through shared memory to form an efficient software pipeline.}
  \label{fig:arch}
\end{figure*}

\subsection{LLM GPU Kernel Techniques: From Local to Global Fusion}
\label{sec:bg-fusion}

GPU operator optimization has evolved from local to global fusion, forming several distinct research directions:

\textbf{Traditional Kernel Fusion} merges adjacent kernels to reduce launch counts, but still relies on global memory for intermediate data exchange, making it difficult to circumvent memory bandwidth and data dependency bottlenecks at extreme model scales.  CUDAGraph reduces launch overhead by capturing and replaying kernel sequences, but cannot achieve deep inter-operator fusion.

\textbf{Compiler-level operator fusion.}  Triton~\cite{triton} achieves compile-time operator fusion through block-level programming abstractions, while Hidet~\cite{hidet} provides finer-grained scheduling control via a task-mapping paradigm.  These approaches improve fusion depth but still operate on a ``kernel-per-operator'' execution model, unable to eliminate operator switching overhead.  Additionally, MCFuser~\cite{mcfuser} achieves high-performance fusion for memory-bound compute-intensive operator chains, Neptune~\cite{neptune} enables advanced fusion of reduction operators by breaking loop dependencies, and Magneto~\cite{magneto} focuses on coordinated optimization of parallel operator structures.

\textbf{Single-operator extreme fusion.}  FlashAttention~\cite{flashattention} fuses the multiple steps of attention into a single kernel through IO-aware tiling, achieving breakthrough memory efficiency; FlashAttention-2~\cite{flashattention2} further optimizes parallelism and work partitioning.  However, these approaches are confined to individual operators.

\textbf{MegaKernel global fusion}~\cite{megakernel} integrates an entire computation block into a single persistent kernel, partitions warps within an SM into producer--consumer roles via Warp Specialization, and achieves pipeline parallelism through multi-level storage and asynchronous copies, thoroughly eliminating operator switching overhead and breaking through global memory access limitations.  MegaKernel represents the furthest advance in the large-kernel direction, achieving full-chain fusion and persistent computation.

\subsection{Analysis of Mainstream Approaches}
\label{sec:bg-mainstream}

Stanford's MegaKernel is built on Tiny-Llama and FlashAttention~\cite{flashattention} primitives.  It decomposes SM resources into five heterogeneous roles---Loader (asynchronous prefetching), Consumer (tensor computation), Storer (asynchronous writeback), Controller (instruction dispatch coordination), and Launcher (resource release management)---leveraging asynchronous cooperation and semaphore synchronization for deep pipeline decoupling of computation, memory access, and instruction dispatch.  Its implementation draws heavily from NVIDIA CUTLASS~\cite{cutlass} Warp-Specialized GEMM designs.  The framework focuses on Hopper/Blackwell hardware and hard-codes tile sizes for specific models (e.g., Llama-1B), lacking support for Qwen variants.  The codebase contains substantial architecture-specific assembly optimizations, making porting to Ada extremely costly.  The initial version supports only low-load Decoder phases, lacking support for long-sequence and large-batch Prefill, leaving a significant gap for industrial deployment.

Mirage MPK~\cite{mpk} is a representative recent work in operator compilation, proposing a tGraph-based multi-level pipeline abstraction that automates the search for tiling strategies and storage allocation.  Mirage~\cite{mirage} further enables cross-level optimization through $\mu$Graph unification, discovering novel optimizations combining algebraic transformations, scheduling transformations, and custom kernel generation.  OLLIE~\cite{ollie} extends the search space of tensor algebraic expressions through derivation-based transformations, Korch~\cite{korch} achieves optimal kernel orchestration via operator fission and constraint optimization, PET~\cite{pet} discovers optimization opportunities invisible to traditional methods through partially equivalent transformations, and TASO~\cite{taso} automates computation graph substitution generation and verification.  These works advance tensor program optimization from different angles, but none addresses MegaKernel-level global operator fusion.  MPK's Managed Pointer mechanism dynamically determines execution paths at runtime, and the resulting if-else branching degrades instruction issue efficiency, preventing it from reaching hardware peak performance.  Its automated search also struggles to converge to global optima within reasonable time for irregular tiles or complex operator chains.

\section{Ada-MK Overall Architecture}
\label{sec:arch}

As illustrated in Figure~\ref{fig:arch}, the Ada-MK system architecture comprises two tightly coupled phases: \emph{Offline MegaKernel Synthesis} (Phase~I) and \emph{Online Warp-Specialized Execution} (Phase~II).

\textbf{Phase~I: Offline MegaKernel Synthesis.}
This phase is responsible for computation graph generation and hardware-constrained automatic search.  The system first parses the logical structure of the Transformer Decoder and LM Head into a fine-grained DAG.  A hardware resource filter then evaluates constraints such as shared memory capacity and register limits, directly pruning invalid branches that would exceed resource budgets.  Within the resulting feasible search space, the system performs profiling to identify the optimal DAG execution trace for the target GPU, and serializes this trace for runtime invocation.

\textbf{Phase~II: Online Warp-Specialized Execution.}
This phase embeds the serialized MegaKernel as a plugin into the TensorRT-LLM inference framework, enabling seamless switching between Prefill (using TensorRT-LLM's native operators) and Decode (using the MegaKernel engine).  Within each streaming multiprocessor (SM), MegaKernel constructs an efficient \emph{warp-specialized software pipeline}: the computational resources are spatially partitioned into multiple warp groups---Loader, Consumer, Storer, Controller, and Launcher---that share SM resources and collaborate through shared memory to maximally overlap memory-access latency with computation.

\section{Core Optimizations}
\label{sec:optimization}

\subsection{Adaptive Shared Memory Management under Resource Constraints}
\label{sec:smem}

\subsubsection{Multi-Dimensional Parameter-Aware Resource Modeling}
\label{sec:smem-model}

We model shared memory allocation from three dimensions---hardware specifications, model architecture, and dynamic workload---enabling fine-grained shared memory management:

\textbf{Hardware-specification awareness.}  From the GPU hardware specifications~\cite{ada_arch}, we first obtain the maximum available shared memory per SM.  We then partition shared memory into instruction pipeline buffers, synchronization metadata, and computation intermediate buffers, and derive the maximum number of available pages:

\begin{equation}
N_{\text{page}} = \frac{\text{SMem}_{\max} - N_{\text{stage}} \times (\text{Instr}_{\text{buf}} + \text{Semaphores} + \text{Scratch})}{\text{Size}_{\text{page}}}
\label{eq:npage}
\end{equation}

\textbf{Model-architecture and dynamic-workload awareness.}  For different model architectures, we compute the shared memory occupied by weights, quantization scales, and other shared data, and compute activation occupancy as a function of batch size.  The pipeline depth ($N_{\text{stage}}$) is adaptively adjusted: on resource-constrained devices we enforce capacity constraints, while on resource-rich devices we increase pipeline depth to maximize throughput:

\begin{equation}
N_{\text{stage}} = \frac{N_{\text{page, total}} - N_{\text{page, weight}} - N_{\text{page, scale}} - N_{\text{page, act}}}{N_{\text{page, per stage}}}
\label{eq:nstage}
\end{equation}

\textbf{K-dimension fine-grained splitting.}  Inspired by a segmented-computation approach, we halve the $K$-dimension tile.  Each loop iteration loads only the weight sub-tile required by the current sub-block, reducing peak shared memory demand by 50\%.

\subsubsection{Shared Memory Page Reuse}
\label{sec:smem-reuse}

The original Stanford MegaKernel implements page reuse across operators (e.g., releasing a page from operator~A for use by operator~B).  We further optimize this reuse strategy.  At execution time, activation pages and weight/output pages are reused based on automated analysis, as detailed in~\S\ref{sec:dag}.

\textbf{Activation--weight page reuse.}  Once activations have been loaded from shared memory into registers, the occupied shared memory can be reclaimed and dynamically reallocated to the Loader role for weight storage.  This increases the number of pipeline stages (pipeline depth) between the Loader and Consumer, effectively hiding memory-access latency while improving instruction-level parallelism.

\textbf{Activation--output page reuse.}  Computed output data typically requires additional shared memory buffering before being written back to global memory.  Since MMA computation depends on registers, the shared memory occupied by activations can be released after copying to registers and reallocated for MMA result storage.

\subsection{Fine-Grained DAG-Based Automatic Search}
\label{sec:dag}

\subsubsection{Foundational Definitions and Modeling}
\label{sec:dag-def}

To precisely characterize MegaKernel's internal execution logic, we define a formal framework based on directed acyclic graphs (DAGs).  Nodes are classified into two categories: data migration (Global$\to$Shared$\to$Register transfers at each level) and computation execution (MMA, dequantization, epilogue), aligned to PTX instruction granularity.  Edges capture two types of constraints: \emph{data dependencies} encoding producer--consumer relationships (a successor can execute only after its predecessor completes), and \emph{resource dependencies} encoding competition for limited shared memory pages (a subsequent load must stall when insufficient pages are available).

\subsubsection{Automated Pipeline Modeling and Search-Path Optimization}
\label{sec:dag-search}

We propose an automated search framework that bridges from low-level IR to pipeline configuration, addressing the limitation that the original MegaKernel lacks fine-grained DAG definitions for efficient pipeline construction.

\paragraph{Step 1: MLIR fine-grained decomposition.}
To overcome the limitation that operator-level dependency representations obscure parallelism opportunities, we leverage Torch-MLIR's multi-level transformation architecture~\cite{mlir}.  Through deep analysis of MLIR dialects and layer-by-layer Lowering~\cite{tensorir}, and inspired by Relax's composable compilation approach for dynamic machine learning~\cite{relax}, we decompose coarse-grained operator black boxes into PTX-level fine-grained dependency traces, providing precise topological support for MegaKernel's static instruction scheduling and resource management, achieving extreme overlap of computation and I/O.

\begin{table}[!t]
\caption{MLIR high-level to low-level primitive mapping.}
\label{tab:mlir}
\scriptsize
\setlength{\tabcolsep}{3pt}
\begin{tabular}{@{}llp{1.8cm}l@{}}
\toprule
\textbf{MLIR High} & \textbf{MLIR Low} & \textbf{Physical Semantics} & \textbf{HW Res.} \\
\midrule
\texttt{memref.copy} & \texttt{nvgpu.device\_async\_copy} & Global$\to$Shared: async prefetch & HBM, SMem \\
\texttt{memref.load} & \texttt{nvgpu.ldmatrix} & Shared$\to$Reg: warp-level load & SMem, Regs \\
\texttt{quant.dcast} & \texttt{custom.dequant} & Dequant: INT4$\to$FP16 & ALU, Regs \\
\texttt{linalg.matmul} & \texttt{nvgpu.mma.sync} & Tensor Core: $16{\times}8{\times}16$ MMA & TC \\
\texttt{linalg.generic} & \texttt{arith.addf} & Epilogue: bias/act fusion & ALU, Regs \\
\texttt{memref.copy} & \texttt{vector.transfer\_write} & Reg$\to$HBM: async writeback & HBM, Regs \\
\bottomrule
\end{tabular}
\end{table}

\paragraph{Step 2: Constructing logical dependencies.}

\begin{figure}[!t]
  \centering
  \includegraphics[width=\columnwidth]{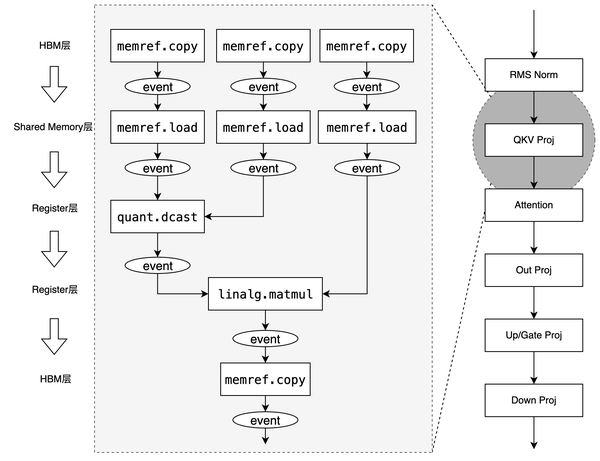}
  \caption{Fine-grained dependency DAG construction from MLIR alias analysis. Write operations identify producers; alias analysis at Read operations determines memory overlap to build RAW edges.}
  \label{fig:dag-construct}
\end{figure}

Based on MLIR's alias analysis capabilities, we automatically identify read-after-write (RAW) dependencies in the instruction sequence, constructing fine-grained dependency DAGs from producers to consumers (Figure~\ref{fig:dag-construct}).  Using MLIR's AliasAnalysis API combined with a \texttt{lastWriters} mapping table, we traverse the instruction sequence, identify memory producers through Write operations, and determine memory overlap at Read operations through alias analysis, thereby automatically constructing RAW dependency DAGs.

\paragraph{Step 3: Schedule and resource candidate-set generation.}

\begin{figure}[!t]
  \centering
  \includegraphics[width=\columnwidth]{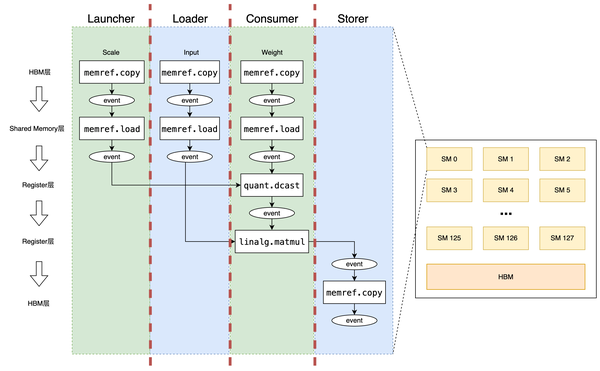}
  \caption{DAG node assignment to four pipeline roles with load balancing and tiling parameter exploration.}
  \label{fig:role-mapping}
\end{figure}

\textbf{Pipeline role splitting (DAG-to-Role Mapping).}  The system assigns DAG nodes to four roles (Figure~\ref{fig:role-mapping})---Launcher (metadata and page scheduling), Loader (asynchronous weight prefetching), Consumer (input loading, dequantization, and MMA computation), and Storer (result writeback synchronization)---and explores mapping schemes of atomic nodes across different roles.  Core strategies include:

\begin{itemize}
  \item \emph{Load balancing:} Avoid stacking ALU-intensive tasks in a single role; migrate lightweight I/O or preprocessing logic to the Loader role to maximize hardware issue slot utilization.
  \item \emph{Tiling parameter space:} Simultaneously traverse multiple Block/Warp tile partitioning ratios to seek the theoretical peak compute throughput.
\end{itemize}

\textbf{Shared-memory-constrained scheduling strategy.}  Shared memory allocation directly determines whether the pipeline suffers from structural stalls.  We define three page states---Empty, Locked, and Ready---identify state conflicts between Loader and Consumer, quantify structural stalls, and compute pipeline duty cycles.  The candidate set is pruned through:

\begin{itemize}
  \item \emph{Early asynchronous prefetching:} Increase the prefetch stride ($N{+}1 \to N{+}2$), trading shared memory space for complete overlap of weight transfer and computation.  TileLink~\cite{tilelink} demonstrates that tile-centric primitives can effectively overlap compute and communication in distributed settings; Ada-MK applies a similar overlapping principle within a single SM's software pipeline.
  \item \emph{Gap filling:} Insert non-critical-path instructions into issue slots left idle by synchronization primitives.
  \item \emph{Address permutation:} Remap logical pages to physical address offsets, reducing bank conflicts through interleaved reuse.  Hexcute~\cite{hexcute} provides constraint-solving references for address permutation, and Axe~\cite{axe} provides a unified layout abstraction for cross-level memory mapping.
  \item \emph{Role rebalancing:} Forward lightweight tasks (e.g., dequantization) to idle Loader roles, accelerating page turnover.
\end{itemize}

\paragraph{Step 4: Static trace search and execution-path solidification.}

\textbf{Heuristic offline search.}  Within the space defined by logical dependencies and physical constraints, we simulate multiple page arrangements and role mappings through offline profiling, locking in the execution trace with the highest theoretical duty cycle.  Unlike Ansor~\cite{ansor}, which employs hierarchical search spaces with evolutionary search, and Pruner~\cite{pruner}, which adopts a Draft-then-Verify search acceleration mechanism, Ada-MK's DAG-level search operates directly on PTX-level dependency graphs, capturing finer-grained parallelism opportunities and effectively handling complex irregular dependencies that traditional MegaKernels cannot address.

\textbf{Execution-path solidification.}  Frameworks such as MPK~\cite{mpk} introduce runtime if-else branches that cause instruction issue delays and pipeline bubbles.  We employ \emph{path solidification} to serialize the optimal execution path directly into kernel code.  This paradigm shift from ``online adaptation'' to ``offline planning'' eliminates the need for runtime resource-state probing, removing the instruction-width overhead of dynamic decision-making.

In summary, this section achieves exhaustive search and locking of physically optimal execution traces in the offline phase through MLIR fine-grained decomposition and DAG dependency modeling, comprehensively considering compute-unit load and storage-page constraints.  This ``offline search, online reuse'' strategy trades compile-time search cost for runtime determinism.

\subsubsection{Overall Effectiveness}
\label{sec:dag-effect}

We systematically analyze the combined performance gains from the fine-grained DAG automatic search mechanism.  Experimental results show that our approach achieves approximately 30\% performance improvement in the Decode phase compared to the original Stanford MegaKernel, with gains arising from two dimensions.

\paragraph{Systematic elimination of spurious dependencies.}
The original MegaKernel implementation contains implicit serialization barriers introduced by static compilation strategies that hinder full exploitation of instruction-level parallelism.  Through fine-grained DAG dependency tracing under data-flow correlation (RAW/WAR) constraints, we achieve two classes of decoupling optimizations:

\begin{figure}[!t]
  \centering
  \includegraphics[width=\columnwidth]{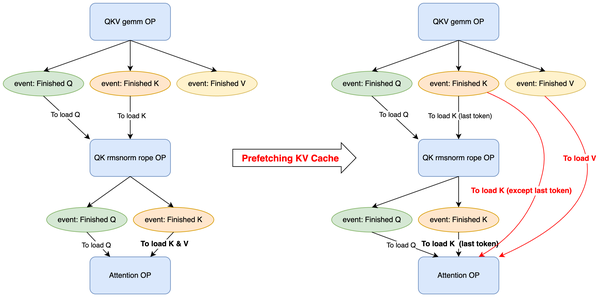}
  \caption{Asynchronous prefetching decouples RMS Norm loads and KV-Cache loads from QKV computation.}
  \label{fig:dep-elim}
\end{figure}

\textbf{Asynchronous prefetching and logical decoupling.}  DAG analysis confirms no data-flow dependency between RMS Norm weights and QKV computation instructions, enabling their load timing to be advanced to the physical page release point, hiding HBM access latency through asynchronous transfer.  Similarly, V-Cache and K-Cache (historical portion) loads in the attention operator are advanced into the pipeline window overlapping with preceding computation, removing historical KV-Cache transfer from the critical path (Figure~\ref{fig:dep-elim}).

\textbf{Pseudo-dependency elimination and streaming reduction.}  In SwiGLU and similar dual-path gated activation structures, conventional implementations trigger Reduce only after both Up/Gate GEMM paths complete, forming a coarse-grained synchronization barrier.  Through fine-grained DAG modeling, we bind the Reduce start condition to the readiness of the corresponding input component, enabling Up-Reduce to proceed in parallel during Gate-GEMM execution, achieving cross-path streaming reduction.

\paragraph{Auto-tuned optimal pipeline configuration.}
Static configuration of warp role allocation and shared memory page scheduling cannot accommodate throughput differences across operators.  Through offline profiling-driven auto-tuning, we achieve configuration convergence in two dimensions:

\begin{figure}[!t]
  \centering
  \includegraphics[width=\columnwidth]{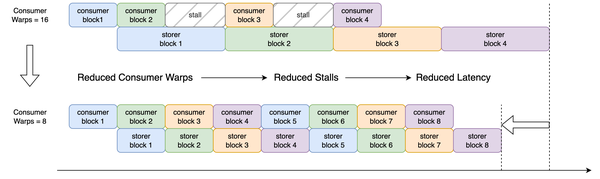}
  \caption{Warp allocation refinement: Consumer warps reduced from 16 to 8 with pipeline stages extended from 2 to 4, aligning role latencies and reducing pipeline stalls.}
  \label{fig:warp-refine}
\end{figure}

\textbf{Warp allocation refinement.}  Reducing Consumer warp count from 16 to 8 (Figure~\ref{fig:warp-refine}) decreases single-GEMM output scale while aligning Storer's cross-block Reduce latency with Consumer's computation latency, effectively mitigating throughput mismatch among pipeline roles.  Additionally, extending the pipeline stage count from 2 to 4 further buffers bubbles caused by execution-latency differences across roles, suppressing pipeline stalls.

\textbf{Page in-place reuse optimization.}  Auto-tuning combined with circular page allocation dynamically identifies shared memory page slots eligible for early release, enabling time-multiplexed reuse of weight and activation pages under WAR constraints, effectively improving the utilization efficiency of the limited shared memory space.

Combining both optimization dimensions, our approach achieves systematic improvement in MegaKernel pipeline efficiency on Ada's resource-constrained architecture through the synergy of dependency decoupling and auto-tuning.

\subsection{Industrial Deployment and End-to-End Optimization}
\label{sec:deploy}

\subsubsection{Heterogeneous Hybrid Inference Engine with Integrated MegaKernel}
\label{sec:hybrid}

MegaKernel demonstrates significant advantages in the Decode phase, but lags behind mature TensorRT-LLM in large-scale parallel computation efficiency during the Prefill phase~\cite{trtllm}.  Three deep-rooted reasons explain this: (1)~\textbf{Decode is IO-bound:} single-token generation involves minimal computation, with the bottleneck in memory access (weight + KV Cache loading); MegaKernel's inter-operator parallelism deeply overlaps computation and I/O for significant gains.  (2)~\textbf{Prefill is compute-bound:} long-sequence large-batch computation saturates GPU compute units, yielding diminishing returns from MegaKernel's IO-compute overlap.  (3)~\textbf{Kernel launch overhead differential:} Decode launches thousands of kernels per token (overhead $\sim$14.6\%), while Prefill executes only one forward pass where launch overhead is negligible, preventing MegaKernel's core advantage from being fully exploited.

Moreover, production environments have accumulated extensive business capabilities built on TensorRT-LLM (e.g., prefix-tree constrained decoding, integrated generation-discrimination).  Completely replacing the TensorRT-LLM engine would incur prohibitive engineering migration costs.  We therefore embed MegaKernel deeply into TensorRT-LLM's execution topology via a Plugin mechanism, constructing a heterogeneous hybrid inference engine that combines the strengths of both:

\begin{itemize}
  \item \emph{Hierarchical operator replacement:} We preserve TensorRT-LLM's mature implementations for peripheral logic (Embedding, KV Cache management) and replace only the core Transformer Block with a custom MegaKernel plugin, minimizing the modification scope.
  \item \emph{Phase-adaptive execution:} The Prefill phase retains TensorRT-LLM's native fused operators, leveraging their high parallelism for large-scale token scenarios; the Decoder phase automatically switches to the MegaKernel engine, eliminating launch overhead and memory-access bottlenecks through full-chain kernel fusion.
  \item \emph{Zero-cost business capability reuse:} Business logic such as prefix-tree constrained decoding and integrated generation-discrimination requires no secondary development for MegaKernel, directly reusing TensorRT-LLM's existing implementations.  This approach simultaneously accommodates Prefill's high throughput and Decode's low latency without additional business refactoring costs, providing a viable path for MegaKernel's large-scale industrial deployment.
\end{itemize}

\subsubsection{MegaKernel Quantization Implementation}
\label{sec:quant}

Building on adaptive shared memory management, we further incorporate quantization optimizations~\cite{qserve} into MegaKernel.  This serves two purposes: quantization compresses weight volume, alleviating Ada's shared memory capacity bottleneck and enabling more pipeline stages; and Ada's inference workloads remain memory-bandwidth-bound, so quantization directly reduces memory pressure and improves throughput.  The implementation comprises three components:

\textbf{TensorCore-aware weight reorder.}  Inspired by QServe~\cite{qserve}, we perform offline weight reordering so that each thread can efficiently access weights during inference.  At runtime, the Consumer role directly loads quantized weights already aligned to TensorCore access patterns via the \texttt{LDMATRIX} instruction, eliminating the need to write dequantized weights back to shared memory and avoiding runtime layout transformation overhead.  Tilus~\cite{tilus} provides reference for layout optimization of arbitrary bit-width low-precision data.

\textbf{K-dimension multi-level pipeline computation.}  Combining K-dimension splitting with quantization, the shared memory required per iteration is compressed from 64\,KB to 32\,KB, reducing page occupancy from 4 to 2.  The freed shared memory capacity is used to extend the pipeline stage count, further deepening compute--memory overlap.

\textbf{Small-batch Padding I/O optimization.}  Under small batch sizes, Tensor Core instructions require fixed tile dimensions, causing significant padding redundancy when effective data is insufficient.  We employ vector-level fine-grained loading with explicit register reorganization to bypass traditional tile-level \texttt{ldmatrix} mechanisms, eliminating redundant I/O of invalid padding data between HBM and registers.  This optimization draws on Tilus~\cite{tilus}'s automatic vectorization and instruction selection strategies, as well as AWQ~\cite{awq}'s activation-aware weight quantization acceleration.

\section{Experimental Evaluation}
\label{sec:experiment}

\subsection{Experimental Objectives and Research Questions}
\label{sec:exp-obj}

We systematically evaluate the performance gains of MegaKernel optimizations within TensorRT-LLM~\cite{trtllm} and compare against mainstream open-source inference frameworks vLLM~\cite{vllm}, SGLang~\cite{sglang}, and vanilla TensorRT-LLM~\cite{trtllm}.  Experiments focus on throughput across different input/output lengths, batch sizes, and task workloads, addressing the following research questions:

\begin{enumerate}
  \item Can MegaKernel significantly improve TensorRT-LLM throughput in low-latency, small-batch inference scenarios?
  \item Does MegaKernel's benefit persist when sequence lengths extend from fixed short sequences to real task datasets?
  \item Does Ada-MK retain advantages over high-throughput frameworks such as vLLM and SGLang at larger batch sizes?
  \item Is MegaKernel's performance benefit consistent across model versions?
\end{enumerate}

\subsection{Experimental Setup}
\label{sec:exp-setup}

\subsubsection{Compared Frameworks}

\begin{table}[!t]
\caption{Compared inference frameworks.}
\label{tab:frameworks}
\small
\begin{tabular}{@{}llp{3.5cm}@{}}
\toprule
\textbf{Framework} & \textbf{Version} & \textbf{Description} \\
\midrule
vLLM~\cite{vllm} & v0.19.0 & High-throughput LLM inference with efficient KV Cache management \\
SGLang~\cite{sglang} & v0.5.10 & Structured generation and high-performance serving \\
TRT-LLM~\cite{trtllm} & v1.1.0rc5 & NVIDIA native inference (baseline) \\
Ada-MK & --- & TRT-LLM + MegaKernel (ours) \\
\bottomrule
\end{tabular}
\end{table}

All frameworks are evaluated in offline batch mode: a fixed batch of requests is submitted simultaneously, and throughput is computed after collecting all generation results.  This mode precisely controls concurrency, eliminating interference from online scheduler differences on throughput measurements, and is suitable for horizontal comparison centered on operator execution efficiency.

\subsubsection{Model and Quantization Configuration}

\begin{table}[!t]
\caption{Model configurations.}
\label{tab:models}
\small
\begin{tabular}{@{}lll@{}}
\toprule
\textbf{Model} & \textbf{Parameters} & \textbf{Quantization} \\
\midrule
Qwen3-1.7B~\cite{qwen3} & 1.7B & GPTQ-W4A16~\cite{gptq} \\
Qwen2.5-1.5B~\cite{qwen25} & 1.5B & GPTQ-W4A16~\cite{gptq} \\
\bottomrule
\end{tabular}
\end{table}

We cover two Qwen-series models of similar scale but different versions to validate the generality of optimization benefits.  Note that the two models have different parameter counts (1.7B vs.\ 1.5B); cross-model comparisons aim to verify MegaKernel's generality under identical quantization, and do not exclude the influence of parameter-count differences on absolute performance.

\subsubsection{Workload Configuration}

Experiments include two workload types:

\begin{enumerate}
  \item \textbf{Fixed short-sequence workload:} Input length 64 tokens, output length 12 tokens (denoted in64/out12).  This scenario evaluates framework throughput under low-latency, short-output generation tasks.
  \item \textbf{Real-dataset workload:} CSL and Human-eval datasets, with context lengths primarily distributed in the $\sim$200--1000 token range, evaluating performance stability under medium-to-long input scenarios.
\end{enumerate}

All experiments are conducted across batch sizes of 1, 2, 4, 8, and 16.  The evaluation metric is generation throughput in tokens/s; higher values indicate better inference performance.

\subsubsection{Experimental Environment}

All experiments are conducted on a single server with an Intel Xeon Platinum 8558 processor (96 cores / 192 threads), 1\,TB DDR5 memory, and a single NVIDIA L20 GPU~\cite{ada_arch} (48\,GB GDDR6, memory bandwidth 864\,GB/s, Compute Capability 8.9).  The OS kernel is Linux 5.10.0, GPU driver 535.161.07, CUDA 12.2.  Each framework runs in an isolated Docker container.  Each experiment exclusively occupied a single GPU, eliminating multi-task resource contention.

\subsection{Fixed Short-Sequence Experiments}
\label{sec:exp-fixed}

\begin{figure*}[!t]
  \centering
  \includegraphics[width=\textwidth]{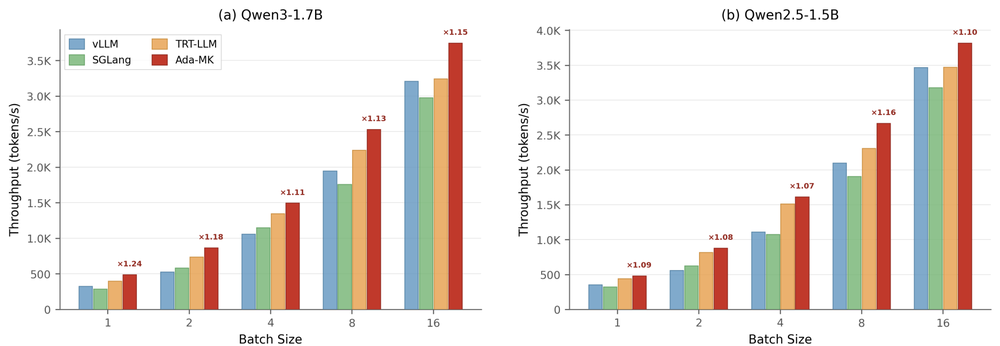}
  \caption{End-to-end throughput comparison on fixed short sequences (input=64, output=12) for (a)~Qwen3-1.7B and (b)~Qwen2.5-1.5B, both quantized with GPTQ-W4A16.  Four inference frameworks are compared: vLLM, SGLang, vanilla TensorRT-LLM, and Ada-MK (ours).  Numbers above the Ada-MK bars indicate speedup over vanilla TensorRT-LLM.  Ada-MK achieves the highest throughput across all batch sizes, with the most significant gains at small batch sizes (up to 23.6\% for Qwen3-1.7B at BS=1).}
  \label{fig:fixed}
\end{figure*}

Under fixed input length 64 and output length 12 (simulating short-input short-output business scenarios), the throughput results across frameworks for Qwen3-1.7B and Qwen2.5-1.5B are shown in Figure~\ref{fig:fixed}.

The results demonstrate that MegaKernel delivers significant benefits in short-sequence output scenarios.  Compared to vanilla TensorRT-LLM, Ada-MK maintains $>$10\% improvement across all batch sizes, with a 23.6\% improvement at BS=1.  This indicates that MegaKernel not only improves large-batch throughput but also significantly reduces execution overhead in small-batch scenarios.  On Qwen2.5-1.5B, Ada-MK similarly achieves the highest throughput across all batch sizes.  Although the improvement over vanilla TensorRT-LLM is slightly lower than on Qwen3-1.7B, it maintains a stable gain of 6.7\%--15.6\%, confirming that MegaKernel's optimization is not limited to a single model version.

\subsection{Real-Dataset Experiments}
\label{sec:exp-real}

\subsubsection{CSL Dataset}

\begin{figure}[!t]
  \centering
  \includegraphics[width=\columnwidth]{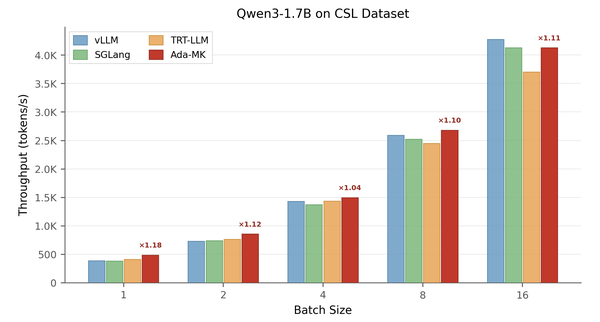}
  \caption{End-to-end throughput comparison on the CSL dataset (Qwen3-1.7B GPTQ-W4A16).  Ada-MK achieves the highest throughput at BS=1 through BS=8, while vLLM surpasses Ada-MK at BS=16 by 3.5\%, indicating that system-level scheduling advantages emerge under high-concurrency long-context scenarios.}
  \label{fig:csl}
\end{figure}

The CSL dataset corresponds to medium-length context inputs, reflecting inference workloads closer to real tasks.  The throughput results for Qwen3-1.7B GPTQ-W4A16 on the CSL dataset are shown in Figure~\ref{fig:csl}.

In the CSL scenario, Ada-MK achieves the best throughput from BS=1 to BS=8, confirming that MegaKernel effectively improves TensorRT-LLM's execution efficiency under medium-to-long input scenarios.  However, at BS=16, vLLM achieves the highest throughput, with Ada-MK lagging vLLM by 3.5\% and essentially matching SGLang.  This phenomenon indicates that as batch size and sequence length increase, vLLM/SGLang's advantages in request scheduling, KV Cache management, or parallel scaling begin to emerge.

\subsubsection{Human-eval Dataset}

\begin{figure}[!t]
  \centering
  \includegraphics[width=\columnwidth]{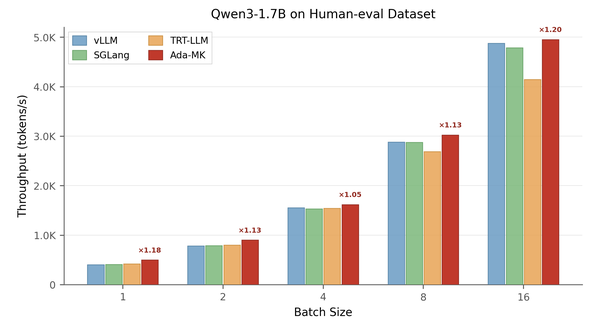}
  \caption{End-to-end throughput comparison on the Human-eval dataset (Qwen3-1.7B GPTQ-W4A16).  Ada-MK maintains the highest throughput even at BS=16, with a 19.5\% improvement over vanilla TensorRT-LLM, demonstrating strong scalability on code-generation workloads.}
  \label{fig:humaneval}
\end{figure}

The Human-eval dataset further evaluates performance in code-generation tasks.  The results for Qwen3-1.7B GPTQ-W4A16 on this dataset are shown in Figure~\ref{fig:humaneval}.

Unlike the CSL scenario, Ada-MK maintains the highest throughput on the Human-eval dataset even at BS=16.  Compared to vLLM and SGLang, its advantage narrows to 1.6\% and 3.5\% respectively at large batch sizes, but it still achieves a significant 19.5\% improvement over vanilla TensorRT-LLM.  This indicates that MegaKernel provides stable improvement to TensorRT-LLM's kernel execution efficiency and maintains good scalability on code-generation workloads.

\subsection{Comprehensive Analysis and Conclusions}
\label{sec:exp-analysis}

\subsubsection{Stable Gains over TensorRT-LLM}

Across all experimental results, Ada-MK achieves positive improvement over vanilla TensorRT-LLM in all tested scenarios and batch sizes.  The improvement reaches up to 23.6\% in fixed short-sequence scenarios and maintains 4.0\%--19.5\% in real-dataset scenarios, demonstrating that MegaKernel's optimization is not specific to a single input pattern or model configuration, but provides universal acceleration for TensorRT-LLM's execution path.

\subsubsection{Most Significant Gains in Small-Batch, Short-Sequence Scenarios}

In BS=1/2 low-batch scenarios, Ada-MK's advantage is most pronounced.  For Qwen3-1.7B's in64/out12 scenario, improvements reach 50.2\% and 64.5\% over vLLM and 71.9\% and 49.3\% over SGLang.  This confirms that MegaKernel effectively reduces kernel scheduling and execution overhead in small-batch inference, making it particularly suitable for low-latency online inference, interactive requests, and short-text generation tasks.

\subsubsection{Narrowing Advantage in Large-Batch, Long-Sequence Scenarios}

As batch size increases and input sequences lengthen, Ada-MK's advantage over vLLM/SGLang gradually narrows.  In the CSL dataset at BS=16, vLLM achieves the highest throughput, with Ada-MK lagging by 3.5\%.  This indicates that in high-concurrency, long-context scenarios, operator/kernel-level fusion optimization alone may be insufficient to fully compensate for differences arising from system-level scheduling and memory management.

\subsubsection{Cross-Model Consistency}

In fixed short-sequence scenarios, both Qwen3-1.7B and Qwen2.5-1.5B show Ada-MK achieving the highest throughput across all batch sizes.  This result demonstrates that MegaKernel's optimization benefits exhibit cross-model consistency and can generalize to different model versions of similar scale and identical quantization.

Overall, MegaKernel significantly improves TensorRT-LLM's inference throughput on GPTQ-W4A16 quantized models, with the most prominent gains in short-sequence, small-batch scenarios.  Compared to vLLM and SGLang, Ada-MK achieves comprehensive leadership in fixed short-sequence scenarios and maintains stable advantages at low-to-medium batch sizes in CSL and Human-eval real-dataset scenarios.  In high-batch-size, medium-to-long sequence scenarios, Ada-MK's advantage narrows, with vLLM/SGLang demonstrating better system-level scalability in some cases.  MegaKernel is an effective optimization for improving TensorRT-LLM inference performance, particularly suited for low-latency, small-batch, short-output online inference tasks.

\section{Conclusion}
\label{sec:conclusion}

This paper addresses the deployment challenges of MegaKernel on resource-constrained NVIDIA Ada architectures and proposes the Ada-MK optimization framework.  Through multi-dimensional adaptive shared memory management, we reduce peak shared memory demand by 50\% and reconstruct efficient pipelines on constrained hardware.  Based on MLIR fine-grained DAG decomposition and offline trace search, we solidify execution paths at compile time, eliminating runtime dynamic decision overhead.  Through a heterogeneous hybrid engine that embeds MegaKernel as a TensorRT-LLM plugin, we simultaneously accommodate Prefill's high throughput and Decode's low latency without additional business refactoring costs.  Experimental results demonstrate that Ada-MK achieves significant end-to-end latency improvements over vLLM~\cite{vllm}, SGLang~\cite{sglang}, and TensorRT-LLM~\cite{trtllm} baselines across multiple batch sizes and task types, representing the first large-scale industrial deployment of MegaKernel in a commercial online advertising system.  Future work will explore Ada-MK's adaptation and migration to larger model scales and next-generation Blackwell architectures.

\bibliographystyle{ACM-Reference-Format}
\bibliography{references}

\end{document}